\documentclass{Interspeech}
\usepackage{siunitx}  

\interspeechcameraready

\title{TalTech Systems for the Interspeech 2025 ML-SUPERB 2.0 Challenge}

\author{Tanel}{Alumäe}
\author{Artem}{Fedorchenko}

\affiliation[nocounter]{Department of Software Science}{Tallinn University of Technology}{Estonia}
\email{tanel.alumae@taltech.ee, artem.fedorchenko@taltech.ee}

\usepackage{comment}

\newcolumntype{P}[1]{>{\centering\arraybackslash}p{#1}}

\begin{document}

\maketitle

\begin{abstract}
This paper describes the language identification and multilingual speech recognition system developed at Tallinn University of Technology for the Interspeech 2025 ML-SUPERB 2.0 Challenge. A hybrid language identification system is used, consisting of a pretrained language embedding model and a light-weight speech recognition model with a shared encoder across languages and language-specific bigram language models. For speech recognition, three models are used, where only a single model is applied for each language, depending on the training data availability and performance on held-out data. The model set consists of a finetuned version of SeamlessM4T, MMS-1B-all with custom language adapters and MMS-zeroshot. The system obtained the top overall score in the challenge.

\end{abstract}

\section{Introduction}

The ML-SUPERB 2.0 Interspeech 2025 Challenge was held with the aim to advance the state-of-the-art in multilingual automatic speech recognition (ASR) and language identification (LID) systems, with a particular focus on improving performance across a diverse range of languages and language varieties. The challenge evaluated systems on 154 unique languages and over 200 language varieties, using a comprehensive scoring system that considers six metrics: average LID accuracy and Character Error Rate (CER) across all languages, average CER for the 15 worst-performing languages, CER standard deviation, and average LID accuracy and CER across language varieties. Participants were allowed to use any available datasets and pretrained models, provided they can perform inference within 24GB of GPU VRAM and without internet connection. This challenge built upon previous SUPERB initiatives \cite{yang2021superbspeechprocessinguniversal} but shifted focus from speech representations to developing robust multilingual ASR and LID systems that ensure no language is "left behind" in speech technology advancement.

The system developed at Tallinn University of Technology (TalTech) for this challenge follows an rule-based pipeline approach: first, a LID model determines the utterance language, then selects the most appropriate ASR model for that language. Since LID accuracy is crucial in this setup, we focused on developing a robust LID system that combines two components: an audio-based language embedding model and a novel generative model. The generative model uses a lightweight multilingual ASR system to identify the language with the highest likelihood decode path, allowing thus to also rely on linguistic constraints. Given the challenge's diverse language set, we primarily used existing multilingual ASR models, developing new models only for languages that were not supported by these models or performed poorly. Our system achieved the highest combined score in the challenge, including the lowest mean ASR CER, the best overall LID accuracy and the highest dialectal LID accuracy.

\section{Methods}

\subsection{Language identification}

We use a hybrid spoken LID model, consisting of a language embedding model with a logistic regression classifier and a generative multilingual speech recognition model.

\subsubsection{Language embeddings model}
\label{sec:lang_emb}

The backbone of the language embedding model is the speech encoder module from the SeamlessM4T speech translation and recognition model\footnote{\url{https://huggingface.co/facebook/seamless-m4t-v2-large}} \cite{communication2023seamlessmultilingualexpressivestreaming}. This speech encoder employs a W2V‑BERT 2.0 model with a Conformer architecture. It was fine‑tuned on 43,772 hours of supervised ASR and speech translation data (as part of the encoder-decoder model), after being pre‑trained on vast amounts of unlabelled speech. This combination enables it to capture a wide range of phonetic and prosodic features across languages. 

The speech encoder module has 24 Conformer layers with 1024-dimensional outputs that are kept frozen during training the language embedding model. The outputs from individual layers are aggregated using dimension-specific weighted average, where the weights specific to each dimension sum to one. That is, the weights are represented by a $1024 \times 24$ matrix. The sequence of 1024-dimensional aggregated speech encodings is then pooled using multi-resolution multi-head attention \cite{wang2020multi}. We used 4 attention heads with two-layer 256-dimensional affine hidden layers. The output from the pooling layer is 8192-dimensional, corresponding to 4 heads and both mean and standard deviation based pooling.
This output is then further processed by two 512-dimensional hidden layers with ReLU nonlinearity and finally by a softmax layer. The model is trained on the VoxLingua107 dataset \cite{valk2021slt} using cross-entropy loss. The model is trained for four epochs, using an effective batch size of 384 utterances, peak learning rate of $10^{-2}$ and weight decay of 0.0001. Reverberation and background noises were used during training, both with a probability of 0.5 per training utterance. We employed the Room Impulse Response and Noise Database from OpenSLR\footnote{\url{https://www.openslr.org/28/}} and used background noises from the MUSAN corpus and simulated impulse responses generated for training ASR models \cite{ko2017study}. After training is complete, 512-dimensional language embeddings can be extracted from the output of the first hidden layer after the pooling layer (before applying the ReLU activation).

In order to train the final language embedding based language classifier, we extracted embeddings for a subset of ML-SUPERB 2.0 training data, augmented with our custom web-scraped data for languages that were not present in the provided data (see Section \ref{sec:data}). This aggregated dataset was filtered to contain up to one hour of speech per language. The final classifier consists of length normalization, LDA with the output dimensionality of 100 and logistic regression.

\subsubsection{Generative language identifier}
\label{gen_lid}

\begin{figure}[t]
  \centering
  \includegraphics[width=\linewidth]{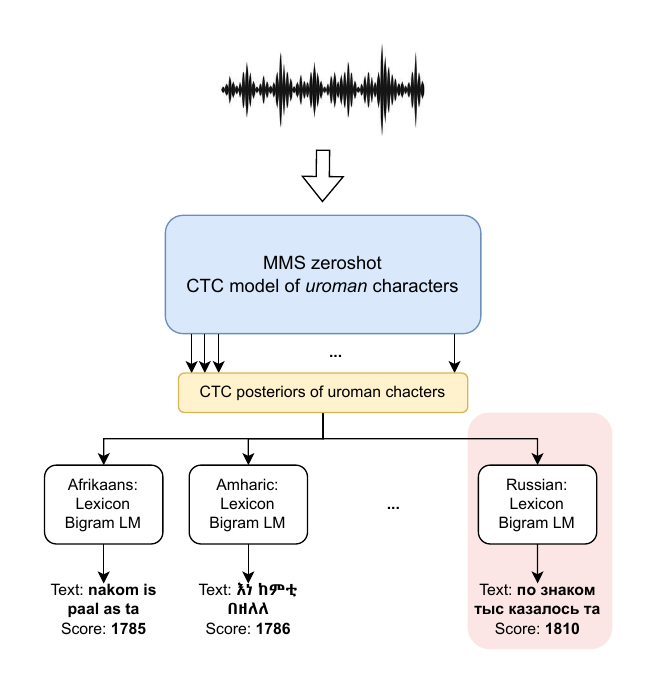}
  \caption{Schematic diagram of the generative LID model based on the MMS-zeroshot model.}
  \label{fig:mms_zeroshot}
\end{figure}

The generative classifier is inspired by phonotactic LID models (e.g., \cite{lamel1994language,zissman1995language} and previous work on accent-robust language classification \cite{kukk2022improving}. Phonotactic spoken LID models work by converting speech into sequences of phonetic units, using several monolingual or one universal phone recognition model. The likelihood of observed phoneme sequences under each language's phonotactic constraints, typically modeled using $N$-gram models that are estimated on the phone transcripts of language-specific training data, is then used to classify the spoken language.

The backbone of our generative language classifier is the recently proposed MMS-zeroshot ASR model \cite{zhao2024scaling}.
Instead of language-independent allophones, the model uses common Latin script characters (known as \textit{uroman} characters) as its intermediate representation. The model trains on labeled data from 1,078 languages, with all labels converted to romanized text (so-called \textit{uroman} characters). Using uroman characters instead of allophones offers a key advantage: mapping low-resource languages to uroman is more reliable than creating word pronunciation lexicons or grapheme-to-phoneme converters, which typically require linguistic expertise. The model uses the uroman toolkit \cite{hermjakob-etal-2018-box} for romanization, which maps characters through heuristics rather than language-specific dictionaries, enabling support for diverse scripts and languages. Also, unlike phonotactic models that need language-specific audio training data to estimate phoneme-based $N$-gram models, the proposed model needs only textual data for each target language for training.

Figure \ref{fig:mms_zeroshot} illustrates our LID model based on MMS-zeroshot. The process begins by passing an utterance through the MMS-zeroshot CTC model, which produces a sequence of posterior probabilities over uroman characters and the blank symbol. These posteriors are then decoded using language-specific decoders, each combining a bigram language model (LM) with a lexicon that maps words to uroman. This generates ASR hypotheses and total decoding scores for each language. We use only the scores, selecting the language with the highest score as the identification result (i.e., Russian on Figure \ref{fig:mms_zeroshot}). When a probability distribution over target languages is needed, we normalize the scores to sum to one.
While this approach requires separate n-gram based decoding for each target language, the computational cost remains reasonable. The CTC posterior computation occurs only once, and the language-specific decoding steps can run in parallel on CPUs using compact bigram LMs. Our implementation  applies this model to an average utterance from the development set in around one second.

Since we use a pretrained MMS-zeroshot ASR model, we don't need any audio data for training this model --  only textual data is needed for building the LMs. For most target languages, we used data from the GlotLID Corpus\footnote{\url{https://huggingface.co/datasets/cis-lmu/glotlid-corpus}} which contains texts for training the GlotLID text-based LID model \cite{kargaran2023glotlid}. For six languages not present in this corpus (Batak, Fulah,  Min Nan Chinese, Southern Thai, Northeastern Thai, Yoloxochitl Mixtec), we scraped data from web sources.

To create the LMs for the LID system, we first sampled up to 100\,000 lines randomly from each language's training data. For each language, we then generated a 10\,000-token vocabulary using the unigram LM sentence-piece approach \cite{kudo2018subword}. These tokens served as the basis for training bigram LMs with Kneser-Ney smoothing. We used the uroman library \cite{hermjakob-etal-2018-box} to create the lexicon that maps these subword tokens to uroman characters.

\begin{table*}[bt]
\caption{Summary of the sources of extra data that was used for training audio-based LID, ASR and LM-based LID models.  }
\label{tab:data}
\centering
\begin{tabular}{@{}l|lll@{}}
\toprule
 & LID & ASR & LM \\ \midrule
Amis & https://klokah.tw & https://klokah.tw &  \\
Sediq & https://klokah.tw & https://klokah.tw &  \\
Balinese & IND-eth  \cite{Sakti2023LeveragingTM} &  &  \\
Batak & IND-eth  \cite{Sakti2023LeveragingTM} &  & https://halomoantondang.blogspot.com \\
Haitian & IARPA-babel201b-v0.2b (LDC2017S03) &  &  \\
Northern Thai & Thai Dialect Corpus \cite{suwanbandit23_interspeech} & \cite{suwanbandit23_interspeech} &  \\
Quechua & Siminchik \cite{cardenas2018siminchik} &  &  \\
Southern Thai & Thai Dialect Corpus \cite{suwanbandit23_interspeech} & \cite{suwanbandit23_interspeech} & https://www.sanook.com \\
Tagalog & IARPA-babel106-v0.2g (LDC2016S13) &  &  \\
Tok Pisin &  IARPA-babel207b-v1.0e (LDC2018S02) &  &  \\
Northeastern Thai & Thai Dialect Corpus \cite{suwanbandit23_interspeech} & \cite{suwanbandit23_interspeech} & \cite{suwanbandit23_interspeech} \\
Fulah &  &  & Leipzig Corpus \cite{goldhahn2012building} \\
Min Nan Chinese &  & CommonVoice & CommonVoice \\
Yoloxochitl Mixtec & OpenSLR 89 \cite{shi2021leveraging} & \cite{shi2021leveraging} & \cite{shi2021leveraging} \\
Sinhala &  & OpenSLR 52 \cite{kjartansson-etal-sltu2018} &  \\
Highland Totonac &  & OpenSLR 107 \cite{berrebbi22_interspeech} &  \\ \bottomrule
\end{tabular}
\end{table*}

\subsection{Speech recognition}

For speech recognition, we relied on three models: SeamlessM4Tv2, MMS-1B-all, and MMS-zeroshot.

\subsubsection{SeamlessM4Tv2}
SeamlessM4Tv2 is a large pretrained speech recognition and translation model using an encoder-decoder architecture. The model supports speech recognition for approximately 100 languages and is primarily trained on web-scraped data containing both audio and corresponding (pseudo-)transcripts, such as subtitles. 

Due to the use of weakly-supervised transcripts during training, SeamlessM4Tv2 doesn't always produce exact transcriptions of speech segments. For example, it might reorder words or replace them with synonyms. To address this limitation and ensure the model follows the challenge data's transcription conventions (such as writing numbers as words), we fine-tuned SeamlessM4Tv2 using the provided training set. We filtered this training data to include only the languages and scripts that SeamlessM4Tv2 supports.

\subsubsection{MMS-1B-all}

This multilingual ASR model is part of Facebook's Massive Multilingual Speech (MMS) project \cite{pratap2023scalingspeechtechnology1000}. The model uses a wav2vec2.0 encoder with 1 billion parameters. The encoder was first pretrained using wav2vec2's self-supervised approach on approximately 500,000 hours of speech data covering over 1,400 languages. The model is finetuned for multilingual CTC-based ASR in two stages: first, a multilingual dataset containing 1,107 languages and 44.7K hours of speech is used to train shared model parameters and a shared output layer. In the second stage, the main model is frozen, the output layer is removed, and only language-specific adapter layers and new output layers are trained for each language.

We employ MMS-1B-all for many languages that SeamlessM4Tv2 either doesn't support or handles poorly. Additionally, we trained new language adapters for several languages and scripts that either lack native support in MMS-1B-all or for which we discovered suitable training data. These languages include Amis, Min Nan Chinese, Sinhala, Southern Thai, Highland Totonac, Sediq, Northeastern Thai, and Yoloxochitl Mixtec.

\subsubsection{MMS-zeroshot}

As described in section \ref{gen_lid}, MMS-zeroshot is a multilingual ASR model that intrinsically produces a CTC probability distribution over uroman characters and can be thus used for decoding any language, given a LM (using language's native characters) and a lexicon that maps LM units to uroman characters.

We used MMS-zeroshot for a few languages that were not supported by neither SeamlessM4T nor MMS-1b-all and for which we could not find labelled ASR training data. For each of those individual languages, we trained a 4-gram LM, using the same vocabulary of 10,000 subwords units as was used for LID.

\section{Data}

\label{sec:data}

The challenge organizers provided the ML-SUPERB 2.0 public set as a baseline dataset for both training and development. This dataset combines various multilingual speech corpora and covers 141 of the 153 target languages. They also provided a development set containing 56 dialects and accents.

For training our audio-based LID model, we needed approximately one hour of unlabeled speech data per language. We primarily used data from the public training set. For the 12 target languages not included in the public set, we collected data from websites, public speech datasets, and IARPA BABEL datasets obtained from the Linguistic Data Consortium (see Table \ref{tab:data}, column 'LID').

To train the LMs for both the generative LID model and MMS-zeroshot based ASR model, we used the GlotLID Corpus, version 3.1. While this corpus contains data for 1941 language/script pairs, six challenge target languages were missing. For these languages, we gathered text data from websites and public web corpora (Table \ref{tab:data}, column 'LM').

Collecting and preparing custom ASR training data is labor-intensive due to the diverse formats of annotated speech data. We gathered additional data for nine languages (Table \ref{tab:data}, column 'ASR'). For Yoloxochitl Mixtec and Highland Totonac, we utilized existing data preparation scripts from the ESPnet toolkit \cite{watanabe2018espnet}. For the Taiwanese indigenous languages Amis and Sediq, we used web-scraping scripts available online\footnote{\url{https://github.com/FormoSpeech/klokah_crawler}}. For the few remaining languages, we developed our own data preparation scripts.

\section{Experimental results}

\subsection{Language identification}

\begin{table}[tb]
\caption{LID accuracies (\%) of the two models and their uniform interpolation on two development sets.}
\label{tab:lid}
\begin{tabular}{l|cP{2cm}P{1cm}}
\toprule
               & Embeddings & Gener. (MMS-zeroshot) & Inter\-polated \\ \midrule
Dev            & 85.3                & 70.7                      & \textbf{89.9}         \\
Dev$_{dialects}$ & 80.5                & 73.2                      & \textbf{84.9}         \\ \bottomrule
\end{tabular}
\end{table}

Table \ref{tab:lid} shows the LID accuracies for both individual models and the uniform linear interpolation of the predicted likelihoods of the individual models. While the language embedding model consistently performs better, combining both models through interpolation yields substantial improvements. This combination reduces the number of errors by 31\% on the regular development set and by 23\% on the development set containing dialectal speech. We experimented with more complex methods for model combination, such as optimizing interpolation parameters (weight and temperature scaler) based on development data but since it provided only a slight improvements over linear interpolation, we opted for the simpler method.

Table \ref{tab:lid-errors} presents the most frequent language confusion pairs in both development sets. Most substitution errors occur between linguistically related languages, as expected. However, two unexpected confusions appear in the dialectal development set: Greek being misidentified as Highland Puebla Nahuatl, and Arabic as Northeastern Thai. Since these language pairs are linguistically distant, these errors suggest that our model hasn't learned robust representations for Highland Puebla Nahuatl and Northeastern Thai. This may be due to the quality of the web-sourced training data we used for these languages. Further analysis is needed to understand the exact causes of these errors.

\begin{table}[tb]
\caption{Most common confusion errors (\textit{reference  $\rightarrow$ predicted}) of the combined LID model.}
\label{tab:lid-errors}
\begin{tabular}{c|c}
\toprule
Dev                                                    & Dev$_{dialects}$                                \\ \midrule
Tamil $\rightarrow$ Telugu                               & German $\rightarrow$ Luxembourgish          \\
Tswana $\rightarrow$ Southern Sotho                      & German $\rightarrow$ Dutch                  \\
Lungga $\rightarrow$ Ganda & Arabic $\rightarrow$ Kabyle                 \\
Northern Frisian $\rightarrow$ Dutch                            & Greek $\rightarrow$ H. P. Nahuatl \\
Urdu $\rightarrow$ Hindi                               & Arabic $\rightarrow$ NE Thai  \\ \bottomrule   
\end{tabular}
\end{table}

\subsection{Speech recognition}

Table \ref{tab:dev_results} shows the ASR CER results across different models, presenting the mean CER averaged over language-specific CERs. For languages not supported by a particular model, we use the CER from the next model up in the ``hierarchy'' when calculating the overall mean. While adding larger models generally improves the average CER, this improvement doesn't apply uniformly across all languages. Therefore, we developed an optimized mapping of languages to models based on development set performance, which we used as our final combined model. This mapping is listed in Table \ref{tab:lid-map}. The last line in Table \ref{tab:dev_results} represents our final system, which uses predicted language labels from our combined LID model and an optimized mix of ASR models, reflecting the evaluation scenario. CER results at the language level are provided in the supplementary material\footnote{\url{https://doi.org/10.5281/zenodo.15550990}}.

\begin{table}[tb]
\caption{CER(\%) results on the two development sets, using different ASR models, using either oracle (\textit{O}) or predicted (\textit{P}) language labels.}
\label{tab:dev_results}
\addtolength{\tabcolsep}{-0.4em}
\begin{tabular}{ll|rr}
\toprule
LID & Model & Dev & Dev$_{dialect}$ \\ \midrule
O & MMS-zeroshot & 22.0 & 24.2 \\
O & + MMS-1b-all (custom) for 133 langs & 9.7 & 17.2 \\
O & \hspace{0.5cm}+ SeamlessM4T (ft.) for 89 langs & 8.2 & 12.7 \\ \midrule
O & Optimized combination & 7.9 & 13.6 \\
P & Optimized combination &  11.7 & 18.5 \\
\bottomrule
\end{tabular}
\end{table}

\begin{table}[tb]
\caption{Mapping of languages to ASR models in our final system.}
\label{tab:lid-map}
\addtolength{\tabcolsep}{-0.4em}
\begin{tabular}{p{\columnwidth}}
\toprule
\textbf{Model}: languages \\
\midrule
\textbf{MMS-zeroshot}: nbl, ssw, tok, tsn, ven \\
\textbf{MMS-1b-all}: abk, amh, ami, asm, ast, azz, bak, ban, bas, bel, bos, bre, btk, ceb, chv, cnh, cym, div, epo, frr, ful, grn, hat, hau, heb, hin, hsb, ibo, ina, kab, kam, kan, kaz, kea, khm, kin, kmr, lga, lin, lit, ltz, lug, luo, mhr, mkd, mri, mrj, msa, myv, nan, nod, nso, nya, oci, orm, pan, pus, que, sah, sin, skr, slk, sna, snd, som, sot, sou, spa, sun, tam, tat, tel, tgk, tgl, tos, tpi, trv, tso, tts, uig, umb, uzb, wol, xho, xty, yor, zul \\
\textbf{SeamlessM4T}: afr, ara, aze, ben, bul, cat, ces, ckb, cmn, dan, deu, ell, eng, est, eus, fas, fin, fra, gle, glg, guj, hrv, hun, hye, ind, isl, ita, jav, jpn, kat, kir, kor, lao, lav, mal, mar, mlt, mon, mya, nep, nld, ori, pol, por, ron, rus, slv, srp, swa, swe, tha, tur, ukr, urd, vie, yue \\
\bottomrule
\end{tabular}
\end{table}

\subsection{Evaluation}

\begin{table}[htbp]
\caption{Results of our system and two organizer baselines on evaluation data. Rankings of our system in each individual category are given in underscores.}
\label{tab:model-comparison}
\centering
\addtolength{\tabcolsep}{-0.5em}
\begin{tabular}
{l|rlrlrlrlrlrlrl}
\toprule
Model &
\multicolumn{2}{p{0.9cm}}{LID$\uparrow$} &
\multicolumn{2}{p{0.9cm}}{CER$\downarrow$} &
\multicolumn{2}{p{1cm}}{STD CER$\downarrow$} &
\multicolumn{2}{p{1.1cm}}{CER 15 worst$\downarrow$} &
\multicolumn{2}{p{1cm}}{Dialect LID$\uparrow$} &
\multicolumn{2}{p{1cm}}{Dialect CER$\downarrow$} \\
\midrule
Baseline 1 & 53.8 &  & 51.9 &  & 19.4 &  & 92.8  &  & 18.8 &  & 74.5 &  \\
Baseline 2 & 74.6 &  & 57.1 &  & 33.5 &  & 118.8 &  &  0.0 &  & 65.9 &  \\
Ours       & \textbf{86.8} & \hspace{-0.2em}$_1$ 
           & \textbf{27.4} & \hspace{-0.2em}$_1$ 
           & 23.9          & \hspace{-0.2em}$_5$ 
           & \textbf{82.5} & \hspace{-0.2em}$_4$ 
           & \textbf{56.6} & \hspace{-0.2em}$_1$ 
           & \textbf{50.0} & \hspace{-0.2em}$_2$  \\
\bottomrule
\end{tabular}
\end{table}

Evaluation was performed on a Dynabench server \cite{kiela2021dynabenchrethinkingbenchmarkingnlp}, using a blind test dataset. In order to run evaluation, teams had to pack the model's logic and all models' parameters to a zip file which would then be uploaded, compiled into a container and run on a server. The uncompressed size of our system was around 28 GB. The evaluation scores of the system, together with those of two organizer baseline systems, are given in Table \ref{tab:model-comparison}.  Our system was ranked as the highest in the evaluation leaderboard, obtaining top scores in three subcategories.

\section{Conclusion}

We presented TalTech's system for the ML-SUPERB 2.0 Challenge, which combines a hybrid LID approach with a hierarchical ASR model selection strategy. The LID system uses both acoustic and linguistic information through the combination of a language embedding model and a generative classifier, achieving 86.8\% accuracy on the evaluation set. For speech recognition, we employed an optimized combination of three pretrained and customized multilingual models, selecting the most appropriate model for each language based on development set performance. This approach resulted in a mean CER of 27.4\% across all languages on the evaluation set.

Future work could focus on developing more efficient ways to leverage linguistic constraints in LID. Additionally, the significant performance variations across languages indicate that developing targeted approaches for the most challenging languages remains an important area for improvement.

\section{Acknowledgments}

This work was supported by the Estonian Centre of Excellence in Artificial Intelligence (EXAI).

\bibliographystyle{IEEEtran}
\bibliography{mybib.rebiber}

\end{document}